\documentclass[10pt,a4paper]{report}
\usepackage{arxiv}
\usepackage[utf8]{inputenc}
\usepackage{amsmath}
\usepackage{amsfonts}
\usepackage{amssymb}
\usepackage{algorithm,algpseudocode}
\usepackage{graphicx}
\usepackage{mathtools}
\usepackage{tikz}
\usepackage{color, colortbl}
\usepackage[colorlinks,linkcolor=blue]{hyperref}
\usepackage[nameinlink,noabbrev]{cleveref}
\usepackage[numbers]{natbib}
\usepackage{setspace}
\author{Amelia E. Pollard\\Department of Computer Science\\University of Manchester\\\texttt{amelia.pollard@postgrad.manchester.ac.uk} \and Jonathan L. Shapiro\\Department of Computer Science\\University of Manchester\\\texttt{jonathan.shapiro@manchester.ac.uk}}
\title{Visual Question Answering as a Multi-task Problem}
\begin{document}
\maketitle
\begin{abstract}
Visual Question Answering(VQA)\citep{antol2015vqa} is a highly complex problem set, relying on many sub-problems to produce reasonable answers. In this paper, we present the hypothesis that Visual Question Answering should be viewed as a multi-task problem\citep{caruana1997multitask}, and provide evidence to support this hypothesis. We demonstrate this by reformatting two commonly used Visual Question Answering datasets, COCO-QA\citep{ren2015exploring} and DAQUAR\citep{malinowski2014nips}, into a multi-task format and train these reformatted datasets on two baseline networks, with one designed specifically to eliminate other possible causes for performance changes as a result of the reformatting. Though the networks demonstrated in this paper do not achieve strongly competitive results, we find that the multi-task approach to Visual Question Answering results in increases in performance of 5-9\% against the single-task formatting, and that the networks reach convergence much faster than in the single-task case. Finally we discuss possible reasons for the observed difference in performance, and perform additional experiments which ruled out causes not associated with the learning of the dataset as a multi-task problem.
 
\end{abstract}
\section{Introduction}
Visual Question Answering (VQA)\cite{antol2015vqa} is the act of showing a machine learning architecture an image of a natural scene and a natural language question about that scene with the expectation of a reasonable or appropriate answer. VQA is a relatively new problem in machine learning research; the first VQA architectures were developed in 2015. Despite being a new research area, it has seen use in numerous applications, including assisting the blind\citep{gurari2018vizwiz}\citep{kollurur2017cogcam}, validating human generated captions on images\citep{lin2016leveraging}, and automatic querying of surveillance video\citep{tu2014joint}. VQA is a highly complex task relying on so many aspects of general intelligence that it has even been proposed as a metric for measuring the general intelligence of Artificial Intelligence systems\citep{zitnick2016measuring}.
 
There are several methods to generate answers for the questions presented by Visual Question Answering, from the simplest yes/no questions\citep{Zhang2016CVPR} to multiple choice\citep{Zhu2016CVPR} to fill-in-the-blank style questions\citep{yu2015visual}. The datasets used in this paper are open-ended, the network may select as its answer any possible answer from the dataset.
 
VQA is also quite different to other machine learning problems and presents its own unique set of challenges. In other machine learning problems, the network need only answer a single fixed question, whereas in VQA, the network must learn to answer multiple unrelated questions, without having previously seen any of the specific questions until runtime. Whilst most machine learning problems will have numerous examples of input and output data for their single fixed question, Visual Question Answering must respond to the more general problem of understanding a scene and the natural language that describes it in order to correctly answer a never before seen question. Thus, it is clear both that VQA is markedly different from other types of machine learning problems and that it involves a great deal of complexity as a research area. 

To fully understand the aforementioned complexities of VQA and how these can be and are addressed, we must look at each problem and the challenges presented by it. There are three main problems that must be tackled by VQA networks, one of which is especially complex. The first of these problems is teaching the network \textit{how to see}. This is handled using computer vision; in this case, we are using a pre-trained deep CNN, VGG19. The second, \textit{how to read}, is tackled by natural language processing, which is handled by a CNN in our research as in \citet{yang2016stacked}. These two areas are entire research fields in themselves and present their own unique challenges which are outside the scope of this paper. 

The third and by far the most difficult problem set is \textit{how to understand}, and it is this problem that our research focuses on. To solve this problem set, we must enable the network to understand the question in the context of the scene and to understand the objects and their relationships in the image. In order to correctly respond to the question and produce a reasonable answer, the network must therefore be trained to understand these problems. This general understanding of natural language and how it applies to an observed scene is hoped to eventually lead to interactive systems of many kinds, from industrial robots capable of being programmed with natural language instructions to manipulate their environment, to smart-home systems capable of interacting with their user in a fluid and intuitive way.

Due to the relationship between these three problems, VQA finds itself on the intersection between several research topics that have seen significant focus in recent years: computer vision, reasoning, and natural language processing. It has therefore received much attention over the last decade. Current state-of-the-art methods rely primarily on deep neural networks, which when designed well, excel at capturing the underlying complexity of input datasets. Many variations on the standard deep neural network have been proposed for VQA problems with varying degrees of success, for example, using Convolutional Neural Networks (CNN) to extract image properties, using Long Short Term Memory (LSTM) or CNNs to process natural language questions, or using attention techniques to reduce dimensionality. Fundamentally, all of these approaches have relied on improving the components of the network for each sub-problem (NLP, image processing, object recognition), and thereby finding their own ways to address one of the three problems presented above.
These approaches to Visual Question Answering have so far fallen into four main categories: 
\begin{itemize}
\item{joint embedding: extracting and then combining features from the image and question. }
\item{attentional models: using features from the question to generate an attentional mask over the image.}
\item{compositional models: in which features from the inputs are used to select pre-trained modules to answer the question.}
\item{and knowledge enhanced models: using an existing knowledge base which can be queried to help answer the question.}

\end{itemize}

Much work \cite{malinowski2015ask}\cite{ren2015exploring}\cite{ma2016learning} has focused on the joint embedding approach. This approach involves combining extracted features of the natural language question with the features of the associated image in a higher dimensional space, before running a classifier on this new high dimensional data. Modifications to this method such as integrating attention methods\cite{yang2016stacked}\cite{xu2016ask} have also been shown to be incredibly effective, resulting in a large quantity of research and current state-of-the-art VQA networks being attention based.\\

However, as VQA is a relatively young research area having been coined in 2015, there remain many research topics which have yet to be explored, especially in the area of reasoning about the inputs, where simple deep neural networks have made up the bulk of this research. Datasets for Visual Question Answering which tackle issues such as bias in the answer sets and provide counter examples for each question have been proposed, though capturing the totality of the domain is a nigh impossible task, and difficulties with truly understanding the datasets still exist. In particular, the complex multi-task nature of the Visual Question Answering datasets has so far been ignored. This is the topic that we tackle in this research. While several VQA dataset authors have taken the time to label the individual task types, no research has been performed to analyse the interaction between those tasks which, while they share the same domain of semantic understanding of visual scenes, focus on very different aspects of that domain. Multiple task types in a shared domain require careful handling in order to prevent destructive interference between those tasks, bringing us to the field of multi-task networks.\\
Multi-task networks are typically designed to solve one primary target task, with one or more additional related tasks also learned during training. These secondary tasks are generally ignored at runtime. Multi-task networks have shown performance increases where the tasks share a domain \citep{caruana1997multitask}. This shared domain biases the hypothesis space into a more generalisable form, which results in increased performance over single task approaches. We hypothesise that the more complex nature of VQA datasets makes them particularly suited for this shared domain utilisation, as each question relies upon a series of subtasks which may be shared between question types.\\

Visual Question Answering datasets like COCO-QA\citep{ren2015exploring} and DAQUAR\citep{malinowski2014nips} contain multiple categories of questions that are easily definable. In COCO-QA, those question categories are labelled by the authors. We propose that the categories of questions (object, number, colour, location) can be thought of as separate tasks in a multi-task environment and further, that the overlap between the tasks (for example, spatial reasoning is required by all the question categories) can be considered as separate tasks also.\\ 

We propose that VQA datasets would be better viewed as a multi-task problem, given that the datasets are composed of several different question types and their corresponding subtasks. With this in mind, measures should be also be taken to avoid the interference problems that occur when training a neural network over multiple separate tasks.

By taking this collection of separate tasks into account when building and training the network, we can avoid common problems inherent in networks where multiple tasks are trained together such as inter-task interference and catastrophic forgetting\citep{french1999catastrophic}.

In the first section of this paper, we perform a simple experiment to demonstrate that VQA datasets can be learned by a multi-task network with a performance increase over a single task network in the simplest case. We then present arguments for alternative explanations for this performance increase. We then demonstrate that that performance increase can be extended to more complex networks, before drawing conclusions and proposing further work.

\section{VQA as a multi-task Problem}
As Caruana\citep{caruana1997multitask} showed in his seminal paper on multi-task learning, a shared domain over multiple tasks results in better generalisation of the internal representation over training just the primary task alone. Such increased generalisation over the domain results in increased performance over single tasks. It is our intention to show that such a generalisation can be attained by training a network by treating the question types in VQA datasets as different tasks.

In this paper, we therefore present the hypothesis that Visual Question Answering datasets can be treated as multi-task datasets, where each question type is a single task within the shared domain of natural scenes. This hypothesis is tested by first testing the simplest possible multi-task formation of the VQA dataset and network, before comparing results against those of a simple single-task version of the network.

\section{Experiments}
\label{experiments}
In order to evidence our hypothesis, we must first show that a performance increase can be obtained in the simplest case by changing from a single task to a multi-task paradigm. For that purpose, we constructed two networks. The first network is a multi-task learning network, where each question type is treated as an individual task in the shared domain. We then created a second network with single-task learning paradigm, as is normally applied in Visual Question Answering networks. These networks were then trained over the same datasets, and performance is compared. We then considered potential sources for the observed performance differences and performed additional tests to rule out those potential sources.
Finally, we reconstructed a network from the visualqa.org leaderboard and constructed an additional variation of this network with modifications to train it in a multi-task manner, to demonstrate that this approach is not an artefact of network architecture.
\subsection{Multi-Task Learning Network}
To allow for the training of multiple question types simultaneously, we constructed a simple multi-task style network with multiple question inputs (one for each question type), a single image input, and multiple answer outputs (corresponding to the question inputs). We utilised a pre-trained ImageNet CNN (VGG19\citep{simonyan2014very}) for image feature extraction and a separate CNN for question feature extraction which was trained by back propagation. A single shared hidden layer then receives all image and question features, before multiple fully connected softmax layers output the predicted answers. All networks are trained until convergence with Nadam\citep{dozat2016incorporating}, before being fine-tuned by SGD with momentum, as described in \citet{keskar2017improving}.

\begin{figure}[h]
  \centering
  \caption{MTL Network Structure. Note that rounded orange rectangles represent sections of the network that are learned during training, while non-rounded blue rectangles indicate fixed weights.}
  \includegraphics[width=0.8\textwidth]{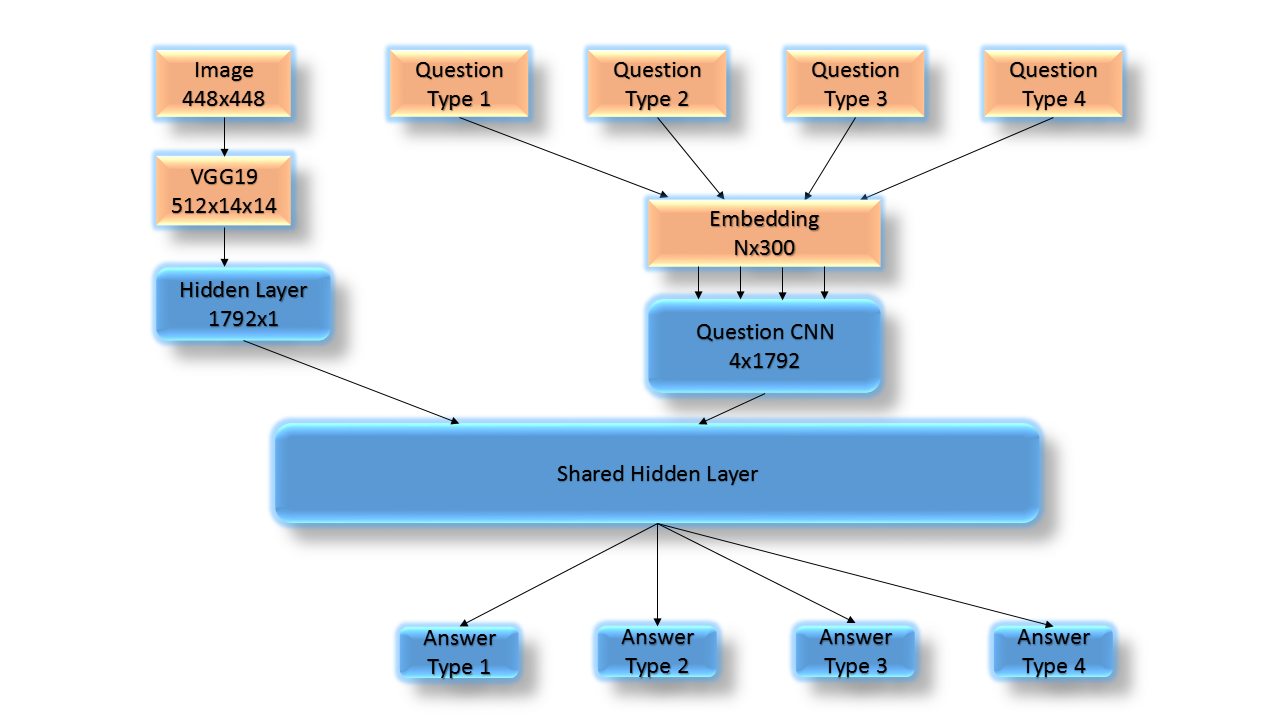}
\end{figure}

The VGG19 CNN is composed of multiple layers of neurons which each complete the desired calculations, before passing onto the next layer of neurons. There are 19 convolutional layers in total, ending with a final classification layer. In this process, we began by extracting the image features from this final classification layer, resulting in a 512 by 14 by 14 matrix of image features. We then compressed these features to a common size by passing them through a linear hidden layer. This resulted in a vector of image features of a much more manageable size.

To generate the question features we used for this experiment, we embedded the questions using GloVe\citep{pennington2014glove} pre-trained embedding, which is then fine-tuned by then end-to-end training of the network in order to increase the training speed of the network. The GloVe database is a publicly available database that was created using a global log bilinear regression model to train a word-word co-occurrence matrix which produces a vector-space with meaningful sub-structure. Essentially, this process results in groups of words with a similar meaning, as high dimensional vectors. The process of embedding converts the natural language words from the questions into a sequence of these high dimensional vectors. Words not present in the GloVe database are initialised with random vectors.
We then passed these embedded questions to the question feature extraction CNN created for this experiment. This question feature extraction CNN is constructed in a convolutional sentence model as described in \citet{hu2014convolutional}. This model uses filters of size 1, 2 and 3, which in this case, would be a single word, a pair of words, and a triplet of words respectively, and applies convolutions to each of these sections to compress the information and extract meaning from the sentences. This results in a feature vector which describes the meaning of each question, both semantically, and syntactically.
Once these two processes are complete, the compressed image features are concatenated with the feature vectors from the question feature extraction CNN, before being fed into the hidden classification layer.

In order to train this network on standard VQA datasets, we must restructure them significantly. Firstly, we require question type labels in order to better utilise the multiple input format. The COCO-QA dataset has question types as part of the label data falling into 4 classes: object recognition questions, numeric questions, colour questions, and spatial questions. The DAQUAR dataset does not have these labels, and so we generated them automatically with keyword searches. Keywords were chosen by the author manually detecting words that were present in only the relevant question types and checked for accuracy by selecting two hundred random samples of questions and the generated type label and verifying them manually; no errors were detected. Some questions defied classification, and were eliminated from the datasets. The majority of these classification failures were due to misspellings in the dataset, though some were due to grammatical errors (for example, "Which object is more?").
The algorithm we designed for automatically classifying question types in the DAQUAR dataset is a simple keyword search. The questions were searched for words associated only with colour questions ('color', 'colour', 'red', 'orange', etc.), positional questions ('on', 'in', 'between', 'left', etc.), numerical questions ('count', 'many', 'number', 'frequent', etc.) and size questions ('largest', 'smallest', 'large', 'big', etc.). It is important to note that the order of the keyword searches is important as some keywords take priority for classifying the question type. For example, "How many orange balls are on the table?" should be classified as a count question as the keyword "How many" takes priority over the keyword "orange". Keywords were therefore searched in the order size, numeric, positional, then colour.

We then collected all images which have multiple question types associated with them and generated training examples by selecting all unique combinations of those questions for each image, padding with zero-vector questions and answers for examples where a question type was not available. The zero-vector placeholder questions were not used for back-propagation or accuracy measurements. For DAQUAR this resulted in a total of 92,288 training examples, while COCO-QA had 240,080 training examples.\\

Finally, we train the network using a sum of the softmax cross entropy across all answer outputs and back-propagation using the entire training set, and test using the entire test set.  

\subsection{Single-Task Learning Network}
To act as a baseline for comparison, we took the reduced datasets from the multi-task pre-processing of the original datasets and reformatted them into the original single task form, with one image and one question associated with one answer. This removed question-answer pairs that were not present in the multi-task training and so prevented the single task network from having the advantage of additional data.\\

To reduce the effect of other possible influences, we preserved as much of the multi-task network's structure as possible by simply removing the additional input and output connections while keeping the hidden layers the same. The final network architecture remained the same as in Figure 1, with input questions 2 to 4 and the corresponding answers removed. As a result, the question feature CNN also has a reduced output size.\\
\subsection{Results}
%add an introductory statement here
\begin{table}[!h]
\centering
\caption{Table of results showing the final accuracy of both the single-task network (STL) and the multi-task-network (MTL) on each dataset, broken down by question type. Performance figures are rounded to one decimal place in accordance with observed variance between runs. The difference between the STL and MTL figures is in accordance with our predictions, with the MTL network consistently outperforming the STL network across all question types.}
\begin{tabular}{c|cccc|c}
\multicolumn{6}{c}{Simple VQA Network Comparison Results} \\ \hline
                     & \textbf{Colour} & \textbf{Count} & \textbf{Position} & \textbf{Size} & \textbf{Total} \\ \hline
                    
\rowcolor[gray]{0.9}
\textbf{MTL COCO-QA} & 25.5\%          & 35.8\%         &  32.5\%           & 28.3\%        & 27.0\% \\
\textbf{STL COCO-QA} & 21.2\%          & 28.4\%         &  27.1\%           & 21.9\%        & 22.9\% \\
\rowcolor[gray]{0.9}
\textbf{Difference}  & +4.3            & +7.4           &  +5.4             & +6.4          & +4.1           \\ \hline
\textbf{MTL DAQUAR}  & 32.0\%          & 23.1\%         &  22.2\%           & 3.4\%         & 17.0\% \\
\rowcolor[gray]{0.9}
\textbf{STL DAQUAR}  & 12.1\%          & 17.4\%         &  0\%\footnote{}   & 2.6\%         & 8.5\%            \\
\textbf{Difference}  & +8.9            & +5.7           &  +22.2            & +0.8          & +8.5        
\end{tabular}
\end{table}

\footnotetext{Due to the unlikely nature of this result we trained the network under a number of different initial conditions, however the result remained that zero of this question type were answered correctly. We are confident in the validity of the result, though we are continuing to investigate.}

The simple nature of the tested network precluded the possibility of state-of-the-art results, however a significant performance increase over the simple STL network can clearly be seen in each individual category and the total result.

The results for the DAQUAR dataset were substantially lower than those of the COCO-QA dataset. We believe this was due to the more complex questions in the DAQUAR dataset, in combination with the much smaller size of the dataset itself which contains only 6794 training examples in contrast with COCO-QA's 78736. The position category questions in particular are often more complex, requiring 2nd order reasoning. For example, "what are on the wall on the left side of the green curtain but not behind the garbage bin".
%add a concluding statement here

\subsection{Alternate Reasons for Performance Changes}
To positively assert that the primary cause of performance changes is the effect of multi-task interference, we must first eliminate other potential explanations. 
Notable possible contributions to performance changes are:
\begin{itemize}
\item Architecture
\item Reduced problem set
\item Shared information
\end{itemize}

\subsection{Architecture}
\label{architecture}
It is possible that the wider input layer allows for more information to be encoded in the connections between the input and hidden layer. With a reduced need to compress the relevant information for all questions types in the hidden layer, it is possible that the multi-task network would have an advantage over the single task network. We test this by training the multi-task network with the single task data, feeding only one question type at a time with blank zero-vectors for the other question type inputs while preserving all examples from the multi-task reformatting. We find that while this does result in a slight performance increase over the single-task network, it does not completely account for the observed increase in performance achieved by the multi-task network trained on the multi-task data.

\begin{table}[!h]
\centering
\caption{Table of results showing the difference in results obtained by testing single task data on the multi-task network in comparison to the results obtained from testing single-task data on the single-task network.}
\begin{tabular}{c|cccc|c}
\multicolumn{6}{c}{Single-Task Data in Both Single and Multi-task Architectures Results} \\ \hline
                     & \textbf{Colour} & \textbf{Count} & \textbf{Position} & \textbf{Size} & \textbf{Total} \\ \hline
                    
\rowcolor[gray]{0.9}
\textbf{STL Data MTL COCO-QA} & 21.3\%         & 28.6\%       &  20.3\%         & 19.3\%       & 22.0\% \\
\textbf{STL Data STL COCO-QA} & 21.2\%        & 28.4\%       &  27.1\%         & 21.9\%      & 22.9\% \\
\rowcolor[gray]{0.9}
\textbf{Difference}           & +0.1         & +0.2        &  -6.8              & -2.6          & +2.9        \\ \hline
\textbf{STL Data MTL DAQUAR}  & 18.2\%        & 30.9\%       &  12.3\%          & 38.6\%       & 9.7\% \\
\rowcolor[gray]{0.9}
\textbf{STL Data STL DAQUAR}  & 12.1\%        & 17.4\%       &  0\%            & 2.6\%       & 8.5\%            \\
\textbf{Difference}           & +5.9         & +13.5        &  +12.3          & +36.0         & +1.2        
\end{tabular}
\end{table}

\subsection{Shared Information}
It is possible that shared information between questions could result in higher performance of the network. For example, the questions "How many oranges are on the table?" and "What fruit is on the table?" share information about the nature of the objects on the table. This could affect the performance in a way that indicates that the tasks are benefiting from a shared domain where in fact they are benefiting from simply regarding the same object. In order to test the contribution of this performance increase we needed to train the network as a multi-task network but test it as a single task network. This demonstrates that the shared representation learned by the network is functioning to represent the domain well, without questions imparting information to each other. Following the training procedure outlined in the section multi-task Network, we then broke from the testing procedure and split the combined questions back into single examples. We tested those questions individually and compared the results to those attained by testing multiple question types simultaneously.\\

We found that the difference in performance was minuscule, on average resulting in a 0.01\% variation, well within the variation observed between runs of the network. This result also enables us to test the network with the original dataset without reductions and thus we are able to present results which are comparable with the results obtained by other researchers.

\subsection{Reduced problem set}
The reduced test dataset could contain less complex examples than the original dataset, leading to an inaccurate perception of increased performance. Given that the result from the investigation into the shared information problem above, this allowed us to test over the complete dataset without reduction, we can rule out an artificially boosted performance score as a potential contribution to our results. However, the reduction in the quantity of training data is likely to reduce overall performance, which does go towards explaining the observed difference between results obtained by our experimentation and those reported by the authors of the network we test in the section "Applicability to other networks".

\subsection{Applicability to other networks}
As we have shown the hypothesised improvement in performance of the multi-task network over the single task network in the simplest case, we then applied the restructured multi-task dataset to a previous competitor for the state of the art. The network chosen was the highest scoring network from VisualQA.org's 2018 challenge leaderboard\citep{vqaorg} which could reasonably be adapted to a multi-task format. Several networks were unsuitable for this task due to the nature of their approach. Specifically, attention based models are particularly problematic for adaptation to a multi-task network format, as this we would require a multi-headed attention implementation which would signifantly increase the computational cost of the network. Due to the current surge in attention based research, this eliminated many potential candidates. We selected "vqateam\_deeper\_LSTM\_Q\_norm\_I"\citep{Lu2015} with an overall accuracy score of 54.08\% as it was the highest scoring network architecture which met the requirement of being sufficiently adaptable. This chosen network performs significantly less well than other networks on the VisualQA leaderboard, however the lack of an attention model makes adapting it to a multi-task architecture much more viable without modifying the network beyond recognition.\\
The vqateam network is composed of the VGG19 CNN for image feature extraction in much the same way as the MTL and STL networks described in \autoref{experiments} (this being a common approach in VQA networks), with natural language processing carried out using a deep LSTM. Both outputs from the CNN and LSTM are then compressed to a common size using a tanh linear layer, before being elementwise multiplied together to combine the features in a manner the authors describe as multimodal. These combined features are then passed through a deep linear network to return a final answer.\\
To modify this network for a multi-task format, we passed all four question types into the LSTM separately, then elementwise multiplied the resulting vectors for each question type with the feature vector extracted from the image. We then concatenated all four vectors and fed the total vector into the deep linear classifier as shown in figure 2.

\begin{figure}[h]
  \centering
  \caption{MTL VQATeam Network Structure. Note that rounded orange rectangles represent sections of the network that are learned during training, while non-rounded blue rectangles indicate fixed weights.}
  \includegraphics[width=0.8\textwidth]{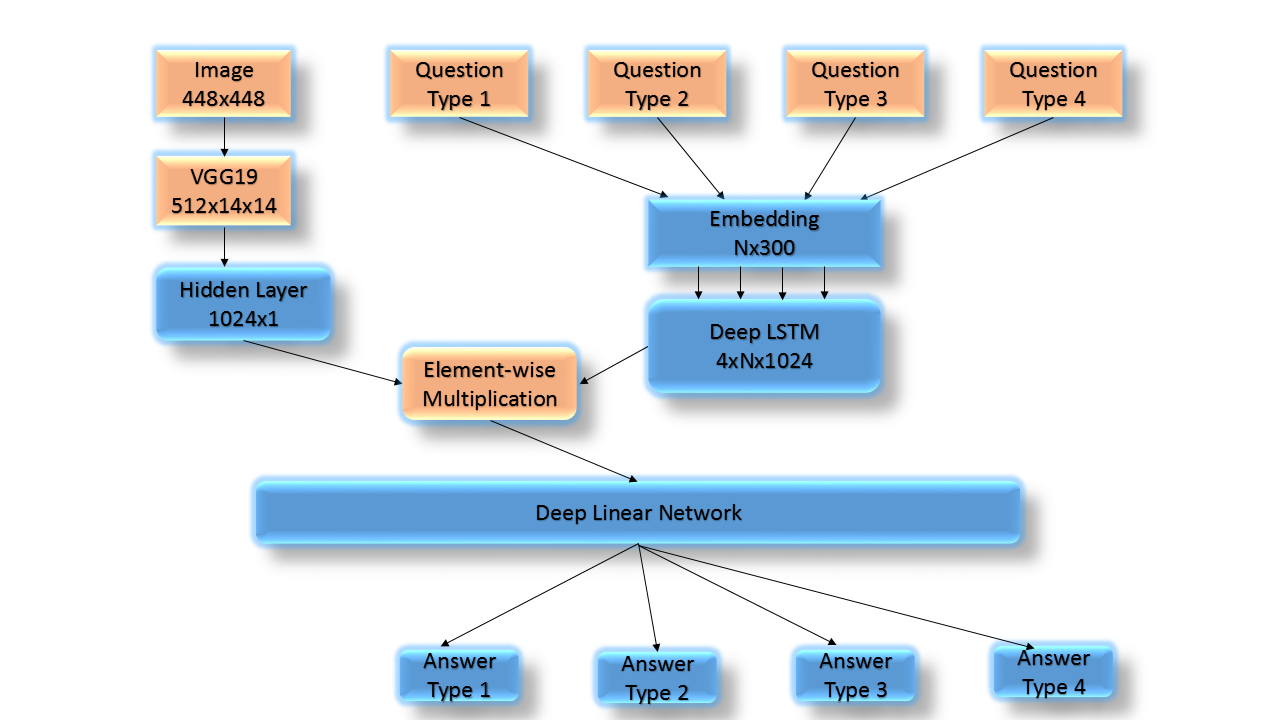}
\end{figure}

As in the experiments performed above, the hyperparameters were found by Bayesian optimisation over a subset of the training dataset and a validation set which was also selected from the training set.\\
The results were as follows:

\begin{table}[!h]
\centering
\caption{Table of results showing the accuracy achieved on the modified VisualQA leaderboard network in both single and multi-task mode, and the corresponding difference between single and multi task performance. }
\begin{tabular}{c|cccc|c}
\multicolumn{6}{c}{Visual QA Leaderboard Network Results} \\ \hline
                     & \textbf{Colour} & \textbf{Count} & \textbf{Position} & \textbf{Size} & \textbf{Total} \\ \hline
\rowcolor[gray]{0.9}
\textbf{MTL COCO-QA}& 2.1\%         & 39.9\%            & 12.0\%         & 0.1\%         &7.1\%                \\
\textbf{STL COCO-QA} &0.1\%         & 1.5\%             & 12.2\%         &0.0\%          &2.5\%                \\
\rowcolor[gray]{0.9}
\textbf{Difference}  &+2.0\%         &+38.4\%           & +12.0\%         & +0.1\%      & +4.6\%               \\ \hline
\textbf{MTL DAQUAR}  & 18.2\%           & 30.8\%         & 12.3\%           & 38.6\%        & 16.3\%         \\
\rowcolor[gray]{0.9}
\textbf{STL DAQUAR}  & 1.6\%           & 34.8\%          & 0.3\%             & 1.3\%        &  10.8\%          \\
\textbf{Difference}  & +17.5\%         & -4.0\%          & +12.0\%           & +37.3\%      &  +6.5\%             
\end{tabular}
\end{table}

In DAQUAR at least, we observe a significant performance increase over the single task networks, further demonstrating the advantage of treating Visual Question Answering as a multi-task problem, though significantly lower overall accuracy than that reported by the authors of the VisualQA leaderboard network. Performance on COCO-QA is very poor in both cases. Although the multi-task network outperforming the single task network still,  neither performed well. We hypothesise that these poor results are the result of the reformatting of the dataset, as the examples that were removed from the datasets are those that do not have multiple questions about the image, and this may be because the image (and thus associated questions) are simpler. Images that contain many objects with many relationships are more likely to have multiple questions about them, while simpler images have fewer questions and so are eliminated from the training set. For reasons which are not clear, this has a significantly greater impact on the VQA Leaderboard network than on the simple network described in \autoref{architecture}. The primary difference between the two networks is the choice of joint-embedding, with the element-wise multiplication of features in the VQA Leaderboard performing much worse in this case than the concatenation of features in the simple baseline network. 
\section{Discussion}
Although the observed performance increase between the single and multi-task formulations of the network was small, it is significant. The elimination of the alternate explanations and the performance increase combined lend considerable evidence to the hypothesis that the contribution of a multi-task learning paradigm to the Visual Question Answering problem results in an improved generalisation over the domain that is not obtained by treating each task in the problem set as a monolithic task. Most notably, the experiment detailed in \autoref{architecture} demonstrates quite soundly that this performance increase is not an artefact of the network architecture and is, in fact, a direct result of training over multiple question types simultaneously.\\
As in other multi-task research, we found that the networks trained in the multi-task format reached convergence much faster than the networks trained in a single-task format, with both multi-task networks achieving their respective final results in less than 50 epochs, while the single-task networks converge only after 200 epochs.
While neither of the networks presented herein achieved state-of-the-art results, the results we attained clearly show the effect of inter-task interference. As a result, we feel confident in stating that a multi-task approach to Visual Question Answering datasets should be considered, and that future work should take into account the distinct nature of each question type as they relate to one another.
\section{Conclusions}
In this paper, we have shown that in the simplest case, Visual Question Answering benefits from being treated as a multi-task problem on both DAQUAR and COCO-QA datasets and show through a process of elimination that this result is in fact due to the increased generalisability of the shared domain. We also demonstrated that this effect is not limited to the simplest case, and while some network architectures are not easily modified into a multi-task architecture, joint-embedding approaches can almost certainly benefit from being so modified. Overall we conclude that the effect of intertask interference should be accounted for in future designs for Visual Question Answering architectures, though ideally in more balanced datasets with many examples per image, as reducing the dataset for such a complex problem has a significant performance impact.\\

\bibliographystyle{plainnat}
\bibliography{vqamtp}

\end{document}